# A novel object slicing based grasp planner for 3D object grasping using underactuated robot gripper

IA Sainul
Advanced Technology Development Centre, IIT Kharagpur,
Kharagpur -721302, India
sainul@iitkgp.ac.in

Sankha Deb
Mechanical Engineering Department, IIT Kharagpur,
Kharagpur -721302, India
sankha.deb@mech.iitkgp.ac.in

AK Deb
Electrical Engineering Department, IIT Kharagpur,
Kharagpur -721302, India
alokkanti@ee.iitkgp.ac.in

*Abstract*—Robotic grasping of arbitrary objects even in completely known environments still remains a challenging problem. Most previously developed algorithms had focused on fingertip grasp, failing to solve the problem even for fully actuated hands/grippers during adaptive/wrapping type of grasps, where each finger makes contact with object at several points. Kinematic closed form solutions are not possible for such an articulated finger which simultaneously reaches several given goal points. This paper, presents a framework for computing best grasp for an underactuated robotic gripper, based on a novel object slicing method. The proposed method quickly find contacts using an object slicing technique and use grasp quality measure to find the best grasp from a pool of grasps. To validate the proposed method, implementation has been done on twenty-four household objects and toys using a two finger underactuated robot gripper. Unlike the many other existing approaches, the proposed approach has several advantages: it can handle objects with complex shapes and sizes; it does not require simplifying the objects into primitive geometric shape; Most importantly, it can be applied on point clouds taken using depth sensor; it takes into account gripper kinematic constraints and generates feasible grasps for both adaptive/enveloping and fingertip types of grasps.

## I. INTRODUCTION

Over the past decades, substantial effort has been devoted in attempt to automate grasping using dexterous multi-fingered robot hand/gripper with a large number of degrees of freedom [1, 2]. Grasping of arbitrary objects even in completely known environment still remains a very challenging problem in robotics. These challenges arise due to several reasons. Firstly, the mathematical complexity of the problem is very high which includes the complex geometry of the objects, nonlinear contact mechanics, incomplete or partial knowledge of the model (such as friction coefficient), etc. Secondly, the constraints posed by robot e.g., unreachability due to hand/gripper kinematic constraints and under-actuation, sensor noise and occlusion in robot perception which leads to grasp failure even in known environments. Methods like form/force [3] closure, in which contact points on the surface of the object are calculated by evaluating grasp quality measure criterions [4]. But the main drawback, by completely ignoring the robot hand/gripper kinematic constraints, is that the contact locations on the surface of the object become unreachable for the real hand/gripper.

Even if there is no reachability problem, the contact positions on the object surface alone are not sufficient to compute the finger joint positions for an under-actuated hand/gripper. Inverse kinematic algorithms fail to solve the problem even for fully actuated hand/gripper during adaptive/wrapping type of grasp where each finger makes contact with an object at several points. It is not possible for an articulated finger to simultaneously reach several given goal points on the object generated by the above grasp synthesis techniques. Thus, it is necessary to compute all the contact points and finger joint positions to successfully perform this type of grasp. Apart from these challenges, the hand/gripper internal degrees of freedom, and those in the wrist create a huge grasp search space. Many approaches are there in the literature to find good wrist position and orientation in this huge search space [5]. Predefined prototypes of grasp have been used to reduce the search space [6]. Hester et al. [7] reduced the size of the grasp search space by assuming fixed wrist position. Vahrenkamp et al. [8] had used principal component analysis (PCA) to align the hand with the object.

In the following, some of the previous works related to the proposed approach are discussed. Rosales et al. [9] introduced the concept of independent contact regions where instead of contact points a finger can make contact anywhere on a region. Xue et al. [10] proposed a swept volume method to find all the contact points between a robotic hand and an object. Saut and Sidobre [11] computed grasp configuration based on an approximation of the intersection between the object surface and the finger workspace and then, selected best grasp according to a quality score. Xue et al. [12] modeled the surface of the objects using super-quadrics and then used continuous collision detection algorithms to find contact points between fingers and the object. Li et al. [13] proposed a novel geometric algorithm to compute enveloping grasp configurations for a multi-fingered hand. The proposed method performs a low-level shape matching by tightly wrapping multiple cords around an object to quickly isolate potential grasping spots. Song et al. [14] voxelized 3D objects and built a contact score map on the object which shows a particular voxel can be touched or not by the fingers/palm.

In this paper, the problem of finding grasp plan for a two-finger, underactuated robotic gripper, based on a novel object slicing method has been discussed. This work is mainly focused on automating robotic grasping of 3D objects of arbitrary shapes and sizes. The 3D object surfaces are represented as polygonal mesh with triangulated facets and vertices. Principal Component Analysis approach has been applied on object vertices to align the gripper with the object. Then, a pool of grasps is generated and the object slicing method is applied to quickly find contact points on the object for evaluating the quality of the grasp. Further, comparison with existing state of the art grasp planning methods has been given to justify relevance and performance of the proposed method in terms of computation time and robustness. Finally, the proposed method has been implemented using a two finger underactuated robot gripper and tested on twenty-four common household objects and toys.

**Contributions:** The important contributions reported in this paper are as follows. The proposed method can handle objects with complex shapes; it does not require simplifying the objects into primitive geometric shapes; with slight changes, it can be applied on point clouds taken using depth sensor; it takes into account gripper kinematic constraints and generates feasible grasps for both adaptive/enveloping and fingertip types of grasps.

The rest of the paper is organized as follows. Section II describes a two finger gripper and its actuation mechanism. Section III presents the grasp planner based on a novel object slicing method. Section IV presents the implementation results and discussions. Finally, Section V gives the conclusions.

## II. UNDERACTUATED ROBOT GRIPPER

In this section, an overview of a two finger underactuated robotic gripper and its actuation mechanism are given. The actuation mechanism of the fingers is discussed in Sainul et al. [15]. The gripper consists of two identical fingers, each having three links namely knuckle, proximal, and distal links as shown in figure 1. Knuckles are fixed on the palm and give support to an object during enveloping/wrapping type of grasp. The knuckle-proximal and proximal-distal links are connected using two revolute joints to form articulated fingers. Tendons run over the pulleys and the routing points inside the fingers and are used to actuate the fingers. Spiral springs and magnetic encoders are used at the joints to open the fingers and measure the joint angular displacements.

A single tendon generates torque at both the joints of the finger. The pulley radii are chosen in such way that torque generated at the proximal joint is greater than the distal joint at equilibrium i.e., $\tau_1 > \tau_2$. On the other hand, keeping spring stiffness values ratio $k_2/k_1 > 1$ makes sure that the proximal joint starts to rotate first when a pulling force is applied to the tendon and stops when any of the following link touches an object or maximum joint limit is reached as shown in Figure 2(a). The distal joint starts to rotate, only once the proximal joint stops as shown in Figure 2(b), where the distal link moves after the proximal link touches the object.

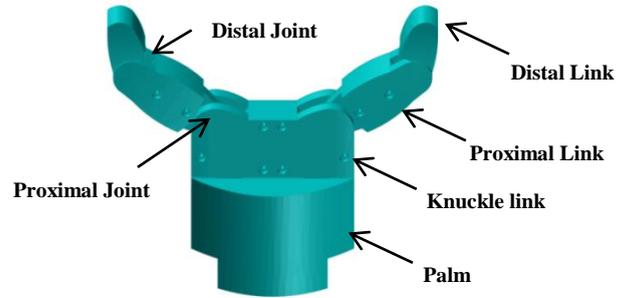

Fig. 1. Model of the robotic gripper

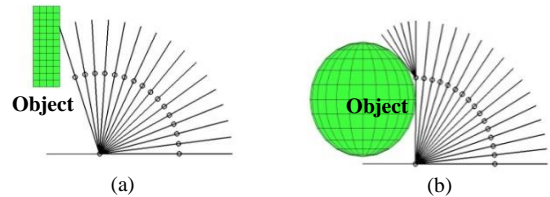

Fig. 2. Finger closure operations

## III. PROPOSED GRASP PLANNER

To start with, the gripper is aligned with the object. Then, a pool of grasps is generated which consists of initial gripper position and approach direction. Then, all the contacts between the fingers and the object are computed for evaluating the quality of the grasps using a grasp quality metric. Finally, the method of close until contacts are found is applied to find the contact points for the best grasp by considering a polygonal mesh for the gripper.

### A. Gripper Alignment

The internal degrees of freedom (DOFs) in the gripper along with 6 DOFs in the robot arm make the grasp space too large to be exhaustively searched. In this subsection, object shape information is used to classify objects into number of categories by applying principal component analysis (PCA) to the vertices of the object. Let the eigenvalues of the PCA be denoted as $\lambda_1, \lambda_2$ and $\lambda_3$ where, $\lambda_1 \geq \lambda_2 \geq \lambda_3$. Then the object is classified into number of categories, for example, principal component of one dimensional object is significantly larger than the other components i.e. $\lambda_1 \gg \lambda_2 \geq \lambda_3$, for two dimensional objects $\lambda_1 \cong \lambda_2 \gg \lambda_3$, and for three dimensional objects $\lambda_1 \cong \lambda_2 \cong \lambda_3$. Further, actual dimensions along the principal components are used to differentiate larger and smaller three-dimensional objects. Then the gripper is aligned with the object according to the type of object fingertips as shown in figure 3, e.g. fingers of the gripper are aligned along the perpendicular direction to the major axis of the cylindrical/bar type of object (one dimensional larger object); fingers are wrapped around a spherical/cuboid object (three dimensional larger object); gripper approaches the plate/flat objects (two dimensional objects) from the edges; gripper grasps a cuboid/box (three dimensional smaller object) using.

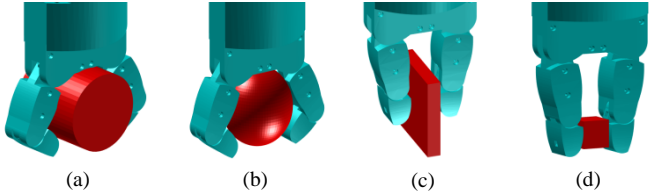

Fig. 3. (a) Cylindrical object grasp, (b) Spherical object grasp, (c) Two dimensional flat object grasp, (d) Smaller object grasp using fingertips.

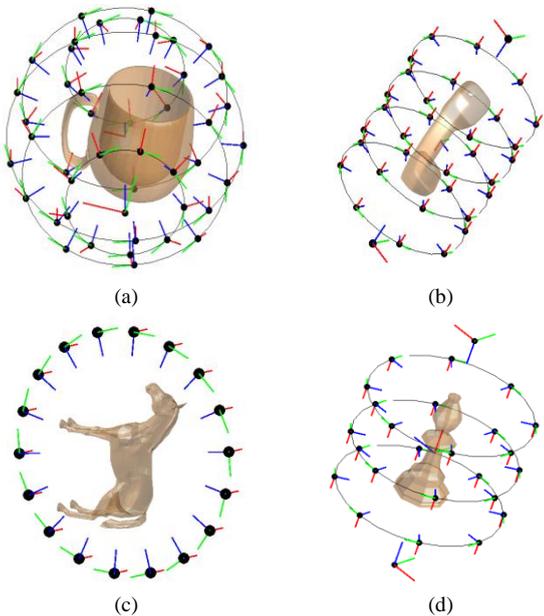

Fig. 4. Different sampling strategies for the generation of grasp pool.

### B. Generation of Grasp Pool

The gripper can approach an object in a large number of ways. In this section, a pool of grasps is generated by placing the gripper at a certain distance from the object and directed towards the object. The number of such grasps is very high, so to keep the number lower a technique similar to Miller et al. [5] is used to generate a pool of grasps by sampling the grasp space. A grasp consists of an initial gripper position, orientation and finger configuration. The orientation gives approach direction of the gripper towards the object. The sampling strategies are predefined for each type of object. For spherical/cuboid object, surface of a sphere enclosing the object is sampled at a fixed angle interval, where the gripper approaches along the radial vector as shown figure 4(a). Cylindrical surface is sampled for spherical/bar type of object, where gripper approaches along the radial vector as shown figure 4(b). An enclosing circle is sampled for flat/plate type of object and the gripper is oriented radially as shown figure 4(c). Both spherical and cylindrical surfaces are used for small box/cuboid type of object with lower sampling rate, where only cylindrical sampling is shown in figure 4(d).

---

**Algorithm 1:** Object slicing based fast contact computation

---

Procedure GraspPointsComputation
    Vertices = ReadObject
    OT = OctreeFormation(Vertices)
    GrsapType = PrincipalComponentAnalysis(Vertices)
    GraspPool =
        GraspPoolGeneration(Vertices, GraspType)
    Loop: For each Grasp in GraspPool
      Contact = ComputeContact(OT, PalmPlane, Palm)
      Loop: For each Finger
        Contact =
            ComputeContact(OT, FingerPlane, Middle)
        If Contact == yes || MaxLimits == Reached
           Contact =
                ComputeContact(OT, FingerPlane, Distal)
      End
    End

Procedure ComputeContacts(OT, Plane, Link)
    Loop: For each Plane
      LeafNodes = Intersection(OT, Plane)
      Points = PointsFrom(OT, LeafNodes)
      ProjectedPoints = Projection(Points, Plane)
      if Link == Palm
        Contact =
            FindNearestPoints(ProjectPoints, Link)
      else
        Contact =
            FindFirstPoints(ProjectedPoints, Link)
    End

A novel object slicing based method is used to quickly find grasp points on the object for evaluating quality of the grasp. The main idea is that a finger does not make contacts with whole surface of an object. If the object is sliced along the plane of finger flexion, then the finger only makes contact with that slice of the object as shown in figure 5(a). The proposed method is performed on vertices of the 3D object model represented as a polygon mesh. It can be also performed on point cloud data of the object taken using a depth camera. In this method, the vertices or points on the cloud are represented in an Octree data structure for fast spatial search. For each grasp in the grasp pool, steps in algorithm 1 given above are applied to compute the possible contacts between the gripper and an object. First the possible contact points with the knuckle are computed. Each grasp from the pool gives the initial gripper position and approach direction relative to the object. Then, the gripper is moved until knuckle makes contacts with the object. To find possible knuckle contacts, a plane perpendicular to knuckle surface and containing the two fingers is chosen. Then, all the leaf nodes of the Octree that are intersecting with the plane are computed and points within these leaf nodes are projected onto the plane. Now, the nearest points are the first to make contact with the palm. For the small objects which cannot be wrapped, the gripper is kept at a fixed distance from the CG of the object. This distance depends on the maximum reach of the fingers. Then, the gripper is positioned with the newly found knuckle contacts. Once the gripper is positioned and aligned with the object, all the points within the leaf nodes that are in intersection with the plane of finger flexion are projected onto that plane as shown in figure 5 (b) and (c). Next, first the proximal and then the distal joints are closed until the proximal and distal links make contacts with the projected points on the plane or maximum joint limits are reached as shown in figure 5(d). Here, the finger links are considered as line segments of actual link length and zero thickness.

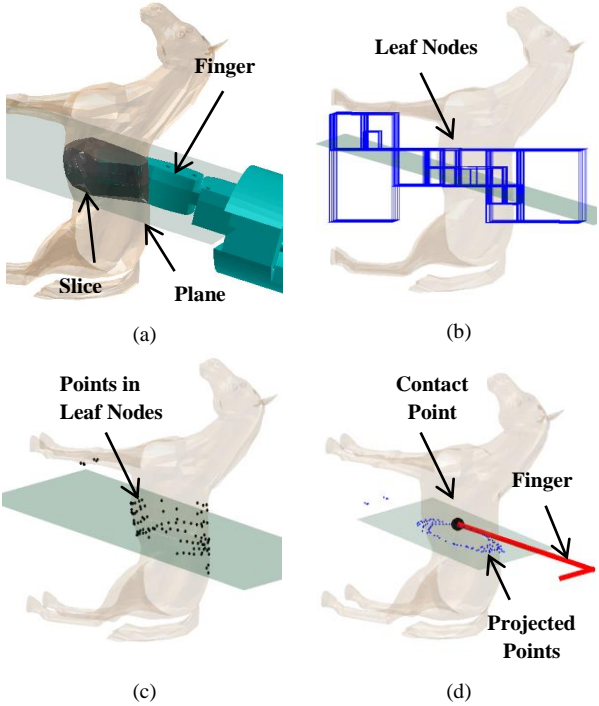

Fig. 5. (a) Object Slicing along the finger plane, (b) Intersection of planes and leaf nodes of the Octree, (c) Vertices of the object inside the leaf nodes, (d) Contact points between the projected points and fingers

*C. Evaluation of Grasp Quality*

Once, all contacts between finger links and object are determined, the grasps are ready to be evaluated for stability using grasp quality measure. Two different quality metrics are combined to form a compact grasp quality metric signifying efficiency of the grasp. The first quality metric determines maximum external disturbance wrench that can be resisted within a unit grasp wrench space by a grasp [16]. The second quality metric tries to grasp the object closest to middle.

The steps involved in finding the first quality metric can be best found in Miller and Allen [17]. Let, n be the total number of point contacts between the links and the object. The contact friction cones at the each contact points are approximated by an eight sided pyramid of unit length. Then the total force $\boldsymbol{f}$ acting on the object at a contact point is the convex combination of eight force vectors approximating the boundary of the cone and must lie within a cone with a half angle of $tan^{-1}\mu_s$, where $\mu_s$ is the coefficient of static friction.

$$\boldsymbol{f} = \sum_j^m \alpha_j \boldsymbol{f}_j \qquad (1)$$

where, $\alpha_j > 0$, $\sum_j^m \alpha_j = 1$ and number of sides of the pyramid $m = 8$

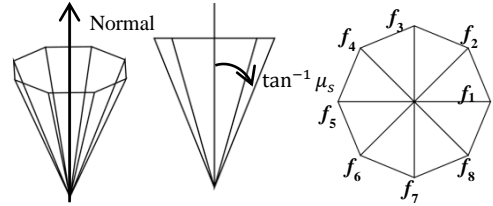

Fig. 6. Approximation of friction cone with an 8 sided pyramid

The wrench space is the set of forces and torques that can be applied by the fingers on the object through point contacts and is given as follows.

$$\boldsymbol{w}_{i,j} = \begin{pmatrix} \boldsymbol{f}_{i,j} \\ \lambda(\boldsymbol{d}_i \times \boldsymbol{f}_{i,j}) \end{pmatrix} \qquad (2)$$

where, $\boldsymbol{f}_{i,j}$ is the $j$-th boundary force vector of the friction cone at the $i$-th point of contact. $\boldsymbol{d}_i$ is the distance from torque origin to the $i$-th point of contact. Multiplier $\lambda$ is chosen to enforce $\|\boldsymbol{\tau}\| \leq \|\boldsymbol{f}\|$.

The convex hull from the wrench space represents the set of wrenches that can be applied on the object given that the sum total of the contact normal forces is one.

$$W = ConvexHull\left(\bigcup_i^n \{\boldsymbol{w}_{i,1}, \boldsymbol{w}_{i,2}, \ldots, \boldsymbol{w}_{i,m}\}\right) \qquad (3)$$

The grasp is stable only if the origin of the wrench space lies within the convex hull. One quality measure that is widely accepted is the distance $\epsilon$ from the origin to the closest facet of the convex hull. This signifies the smallest maximum wrench that can be exerted by the grasp. The scale of $\epsilon$ varies from 0 to 1, closer to 1 signifies more efficient grasp. However, this measure is not invariant to choice of torque origin. The volume $v$ of the convex hull can be used as an invariant quality measure.

Intuitively, on a daily basis if the object is grasped at the middle, then it becomes easier to hold the object. Torque induced by the gravity becomes greater as the objects are grasped further from the centre of gravity. So, the second quality metric $d$ is defined as the distance between the origin of wrench space and the centre of gravity. The normalized quality metrics $\epsilon$ or $v$, and $d$ are combined to form a compact grasp quality metric as follows.

$$Q = \frac{1}{(1+d)} + x \quad (4)$$

where, $x = \begin{cases} \epsilon, & \text{distance measure of the convex hull} \\ v, & \text{volumn measure of the convex hull} \end{cases}$

The above quality measure is used to rank all the grasps in the grasp pool and the best grasp is chosen for an object.

### D. Computation of Contacts and Joint Displacements

In previous sub-section, contacts are quickly computed for finding quality of the grasp by considering each link as a line segment having actual link length and zero thickness. Once the best grasp is found then, a mesh-mesh intersection technique is used to find the contacts, where the object and the gripper links are represented as triangulated meshes. The joint displacements are computed by closing the fingers until contacts are found.

## IV. RESULTS AND DISCUSSIONS

A total of twenty-four common household objects and toys have been chosen for performing grasping using the two fingered gripper. Eighteen objects are taken from the Princeton Shape Benchmark [18] and six objects are designed. Figure 7 shows the different stages of contact computation between an object and a finger. The implementation has been done using Matlab platform running on a PC with configuration Intel Core i3 5005U CPU at 2.00 GHz, 4GB RAM. To implement the proposed grasp planning algorithm, the surface of the object is modeled using triangulated facets and vertices. The algorithm first imports all the facets and vertices data from the 3D CAD files of the respective objects. Here, only results of a subset of six objects with diverse shapes and sizes from the total of 24 objects are given. All the generated grasps for each object are tested using the grasp quality measure as given in table 1. Found grasps is the number of grasps found from the pool of grasps having good grasp quality. Time per grasp is the average computation time (in seconds) to get a good grasp which is the ratio of the total run time to the total number of good grasps.

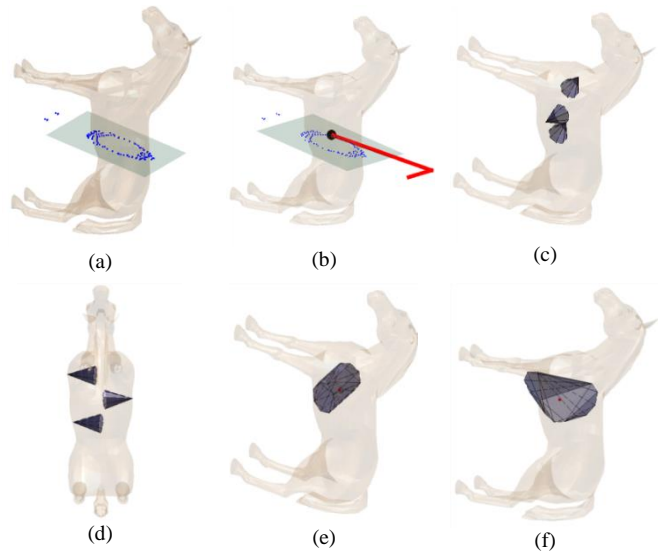

Fig. 7. (a) Vertices projected on the finger plane, (b) Finger makes contact with the projected points, (c) & (d) Contact friction cones with approximated eight sides, (e) & (f) Convex hull for contact force and torque components, respectively.

The grasp planner presented by Li et al. [13] has been implemented for comparison with the proposed grasp planner. Although multiple cords have been used in method proposed by Li et al., here for simplicity, only a single cord has been used and the computation time per grasp for various object is given in table 1. Three dimensional Octree has been used for implementing both the planners to speed up the spatial search. On comparing the time per grasp for both the methods, it is found that the proposed method is nearly 15-70 times faster than the previous one depending upon the types of grasp. Unlike the work presented by Li et al., which is based on ray tracing techniques, the proposed planner works successfully even on objects with small holes on the surface as it does not need facet information. Further, a single cord around the region is not enough to capture whether the region is good for grasping. To overcome this, Li et al. had used multiple cords which, however, give rise to increase in computation time. But in the method that is presented in this paper, the entire contact region is explored to find possible contacts by projecting on the finger plane in a single step, which is computationally more efficient. Other advantages offered by the proposed planner is that it does not depend on the underlying representation of the object model and only needs vertices data, and so it can be directly applied on point clouds taken using depth sensor without any expensive pre-processing such as surface reconstruction. This makes the proposed planner robust as it can handle different object representations e.g. polygonal mesh, unstructured point clouds, etc. Figure 8 illustrates grasping of various objects using the best grasp strategy from the pool of grasps.

One limitation of the proposed method is that it fails on objects having sparse vertices, to handle such objects surface of the objects are up-sampled to increase the data points.

Table 1 Results of implementation of proposed object slicing based planner and comparison with the cord wrapping based planner.

| Objects | Object slicing based planner | | | | Cord wrapping based planner (Li et. al.) |
|---|---|---|---|---|---|
| | Tested grasps | Found grasps | Total time | Time per grasp | Time per grasp |
| Bottle | 216 | 119 | 14.20s | 0.1176s | 3.2558s |
| Apple | 180 | 170 | 17.56s | 0.1033s | 7.8293s |
| Horse | 19 | 15 | 1.78s | 0.1191s | 2.9182s |
| Phone | 144 | 84 | 22.58s | 0.2689s | 6.1769s |
| Mug | 180 | 102 | 11.07s | 0.1153s | 5.6721s |
| Toy Car | 45 | 30 | 3.37s | 0.1126s | 1.6257s |

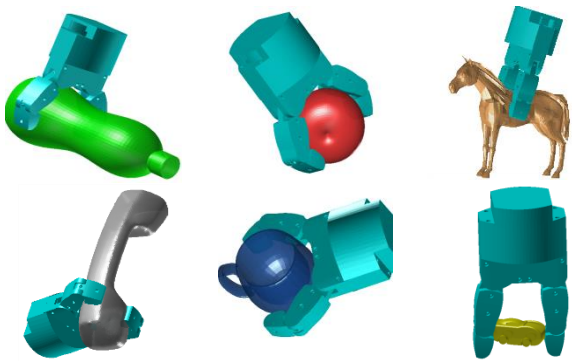

Fig. 8. Final results of grasping of various objects using the best grasp strategy from the pool of grasps

## V. CONCLUSIONS

In this paper, the problem of finding grasp plan based on a novel object slicing based method has been developed and implemented using a two finger underactuated robot gripper. It has been tested on common household objects and toys with different geometric shapes and sizes. The average computation time per grasp is found to remain nearly the same for all the objects. Further, the proposed method is found to be about 15-70 times faster than the method proposed by Li et al. In an attempt to accomplish automatic grasping, the following are some important findings of the work.

i. A novel object slicing based grasp planner has been developed which can perform both adaptive/enveloping and fingertip types of grasps.

ii. Unlike many other approaches, the proposed approach has several advantages: it can handle objects with complex shapes and sizes; it does not require simplifying the objects into primitive geometric shapes; with slight changes, it can be applied on point clouds taken using depth sensor; it takes into account gripper kinematic constraints and generates feasible grasps for both adaptive/enveloping and fingertip types of grasps

In future, the proposed planner will be implemented on depth sensor data. Collision avoidance module will be added to avoid any collision with the environment.